\documentclass{article}

\usepackage{arxiv}

\usepackage[utf8]{inputenc} 
\usepackage[T1]{fontenc}    
\usepackage{hyperref}       
\hypersetup{hidelinks}
\usepackage{url}            
\usepackage{booktabs}       
\usepackage{amsfonts}       
\usepackage{nicefrac}       
\usepackage{microtype}      
\usepackage{lipsum}
\usepackage{graphicx}
\usepackage{natbib}
\usepackage{doi}
\usepackage{supertabular}
\usepackage{afterpage}
\usepackage{multicol}
\usepackage{multirow}
\usepackage{booktabs}

\usepackage{xurl} 

\title{On the Limitations of Large Language Models for Conceptual Database Modeling}

\author{
\vspace{1em}
\href{https://orcid.org/0009-0006-2828-1969}{\includegraphics[height=1.2ex]{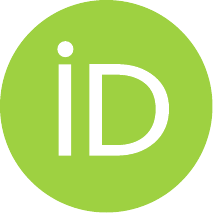} \textbf{Arthur Félix}}\textsuperscript{1} \quad
\href{https://orcid.org/0009-0006-7766-9644}{\includegraphics[height=1.2ex]{orcid.pdf} \textbf{Carlos D. S. Nogueira}}\textsuperscript{1} \quad
\href{https://orcid.org/0009-0002-4610-7092}{\includegraphics[height=1.2ex]{orcid.pdf} \textbf{Eduarda Farias}}\textsuperscript{1} \\
\href{https://orcid.org/0000-0003-4404-2344}{\includegraphics[height=1.2ex]{orcid.pdf} \textbf{Claudio E. C. Campelo}}\textsuperscript{1} \quad
\textbf{Júlia Menezes}\textsuperscript{1}
\\[1em]
\textsuperscript{1}Systems and Computing Department\\
Federal University of Campina Grande (UFCG)\\
Campina Grande -- PB -- Brazil
\\[2em]
\small
\url{arthur.felix.siqueira@ccc.ufcg.edu.br}, 
\url{carlos.daniel.silva.nogueira@ccc.ufcg.edu.br}, \\
\url{maria.eduarda.farias@ccc.ufcg.edu.br}, 
\url{campelo@dsc.ufcg.edu.br}, \\
\url{julia.menezes@ccc.ufcg.edu.br}
}

\date{}

\renewcommand{\undertitle}{Preprint. Under review at Journal of Information and Data Management (JIDM).}
\renewcommand{\shorttitle}{\textit{LLMs for Conceptual Database Modeling}}

\hypersetup{
pdftitle={On the Limitations of Large Language Models for Conceptual Database Modeling},
pdfsubject={cs.AI},
pdfauthor={Arthur Félix, Carlos D. S. Nogueira, Eduarda Farias, Claudio E. C. Campelo, Júlia Menezes},
pdfkeywords={Conceptual-to-Logical Modeling, Database Modeling, Large Language Models, Logical Modeling Errors, Prompt Engineering, Diagram Validation}
}

\begin{document}
\maketitle

\begin{abstract}
Large Language Models (LLMs) have emerged as promising tools for transforming natural language requirements into database models, but their reliability for conceptual-to-logical database modeling remains unclear. This study investigates the effectiveness of LLMs in generating logical relational models from textual requirements under different prompting strategies and increasing levels of modeling complexity. Two LLMs, Qwen-3 and GPT-5.1, were evaluated using One-shot, Chain-of-Thought (CoT), and Chain-of-Thought with Verifier prompting strategies within a reproducible pipeline that transforms model outputs into visual logical relational diagrams using ERDot. The evaluation combines qualitative assessment with a quantitative Checklist Quality Score (CQS), covering completeness, correctness, clarity, simplicity, flexibility, and implementability, as well as an analysis of token usage and inference cost. The results show that GPT-5.1 achieved the highest overall quality, particularly with CoT + Verifier, while Qwen-3 exhibited stronger tendencies toward under-specification. Nevertheless, all evaluated configurations presented substantial deficiencies in requirement coverage, and none reached the level of structural robustness and data integrity defined in the evaluation framework. Moreover, the additional reasoning and verification stages substantially increased computational cost without eliminating the main modeling deficiencies. These findings indicate that, although LLMs can provide useful assistance in database modeling, their current reliability and cost-quality trade-offs do not support autonomous end-to-end generation of logical relational models, reinforcing the need for human validation.
\end{abstract}

\noindent\textbf{Note:} This paper is an extended version of a work originally published in the \textit{Proceedings of the 40th Brazilian Symposium on Databases (SBBD 2025), Fortaleza, CE, Brazil \citep{menezes2025database}}.


\section{Introduction}
\label{sec:intro}

Relational databases are grounded in the relational model, in which data are organized into relations, usually represented as tables \citep{elmasri2011fundamentals}. This model remains widely adopted due to its maturity, ease of maintenance, and structured access through query languages such as SQL (Structured Query Language) \citep{robison2024benchmark}. The design of a relational database requires the careful definition of a logical schema derived from system requirements, since this schema establishes the conceptual and operational structure of data storage.

In this context, diagrams that describe entities and their relationships play a central role in conceptual modeling. The Entity–Relationship model, proposed by \citep{chen1976}, provides an abstract representation of the domain, in which entities correspond to relevant objects or concepts and relationships express interactions among them, often qualified by cardinalities \citep{magalhaes2010}.

Despite their visual clarity, the construction of ER diagrams is highly dependent on the correct interpretation of the requirements, which frequently exhibit ambiguities, omissions, and inconsistencies. Conceptual modeling therefore constitutes one of the most challenging stages of database design, requiring critical analysis, iterative validation, and discussion among specialists. Even when automated tools are employed to generate graphical representations, the quality of the model remains conditioned on the soundness of the underlying conceptual abstraction \citep{bagui2003erd}.

Although there are well-established tools for drawing ER diagrams, most of them operate only at the representation stage, placing full responsibility on the analyst for identifying entities, attributes, and relationships. More recently, LLMs have been explored as potential aids to modeling based on textual requirements written in natural language. However, the degree of reliability of these approaches, especially in more complex requirements scenarios, is not yet fully understood.

The performance of LLMs in tasks that require structured inference depends significantly on how instructions are formulated, making prompt engineering a central element in this process \citep{sahoo2024systematic}. More sophisticated prompting techniques seek to explicitly guide the model’s reasoning, reducing ambiguities and improving response consistency \citep{bansal2024prompt}. Among these techniques, Chain-of-Thought (CoT) stands out by encouraging models to generate intermediate reasoning steps before producing a final answer, a strategy shown to substantially improve their ability to perform complex reasoning tasks \citep{wei2022cot}.

Complementarily, verifier-based approaches introduce mechanisms for validating the generated responses, selecting those that exhibit greater logical coherence and correctness. These strategies highlight that both the prompt structure and the verification mechanisms exert a direct influence on the quality of the generated outputs, especially in tasks involving conceptual abstraction, such as logical schema generated from the conceptual interpretation.

The present study constitutes a continuation of the work ``Database Modeling Automation from Natural Language Requirements'' \citep{menezes2025database}, further deepening the investigation into how different prompting techniques and varying levels of complexity in system requirement descriptions affect the performance of language models. By extending this effort, the study aims to systematically understand the limitations and potentialities of LLMs when applied to conceptual-to-logical database modeling.

The contributions of this work are:

\begin{itemize}
\item A synthesized literature review on the use of LLMs to support conceptual database modeling.
\item An exploratory experimental approach for the automatic generation of logical relational diagrams generated through ERDot from textual requirements.
\item A comparative analysis of different LLMs, evaluating their behavior and limitations in the conceptual interpretation task.
\item A comparative analysis of different prompting techniques, empirically investigating how distinct prompt formulation strategies impact the quality, completeness, and conceptual consistency of the generated diagrams.
\item The construction of a novel dataset containing textual requirements and reference logical relational diagram, aimed at the systematic evaluation of these approaches.
\end{itemize}
 
\section{Related Work}
This section presents the main concepts and approaches related to the automatic generation of diagrams, with an emphasis on both classical Natural Language Processing (NLP) methods and emerging approaches based on Large Language Models (LLMs).

\subsection{Automatic Generation of ER Diagrams from Natural Language Using Classical
NLP Methods}
The automatic generation of Entity-Relationship diagrams using classical NLP methods is a topic already explored in the literature. \citep{btoush2015generating} propose an approach to the English language that employs preprocessing techniques such as tokenization, \emph{part-of-speech} (POS) tagging, \emph{chunking}, and parsing, combined with heuristics to identify entities, relationships, and attributes. Similarly, \citep{indonesianRef} use classical NLP techniques along with a rule-based methodology to identify the elements present in ER diagrams from specifications in the Indonesian language.

In the context of the Vietnamese language, \citep{vietnameseRef} propose a multi-phase methodology that initially uses paraphrasing techniques and heuristic rules to identify diagram elements, and then applies additional heuristics to refine the labeling of these components.

Despite their successes, rule-based ER extraction methods are inherently brittle and language-specific. They depend on manually defined grammars or lexicons tailored to a specific language or domain. For instance, the hybrid system proposed by \citep{Vyramuthu2025} for extracting ER models from German texts still relies on a rule-based extraction core and explicitly reports difficulties in handling contextual information, ambiguity, and complex ER constructs such as multi-valued attributes and weak entities. In practice, supporting a new language or adapting to a different writing style requires redesigning many of these rules. This rigidity limits portability; for example, a tool built with German or Vietnamese syntax rules cannot easily process Portuguese or Chinese texts without rewriting its rule set.

Although these approaches have made significant contributions to the field, they rely on rigid techniques dependent on specific linguistic rules, which limits their scalability and adaptability to other languages or domains. In contrast, the present work proposes a new methodology based on LLMs for the automatic generation of logical relational diagram components. This approach stands out not only for employing modern natural language processing techniques but also for its flexibility: it is not restricted to a single language and can be applied to any language with operational LLMs. This characteristic significantly expands the reach and applicability of the proposed solution.

\subsection{Use of LLMs for Diagram Generation}
The use of LLMs for diagram generation constitutes an emerging area of study. Several works have initially demonstrated the generation of general-purpose diagrams from natural language texts. \citep{diagrammerGPT2023} propose a \textit{framework} that enables the construction of a new type of diagram, an diagram plan, that can encompass across various domains -- both within computing and in external fields — from textual descriptions. Furthermore, \citep{umlPrompt2023} suggest an approach for creating UML diagrams using \textit{prompt engineering} techniques. These studies provide a relevant methodological foundation for the present work, which specifically focuses on the application of LLMs in the automatic generation of Entity-Relationship (ER) diagrams. However, although these approaches have shown promising results in various contexts, they have not been directly evaluated for ER diagram generation. This gap indicates that the potential of these methodologies for this specific type of diagram remains unknown, with potential limitations or challenges that have yet to be explored.

Regarding the generation of ER diagrams, the application of LLMs is still in its early stages. Although \citep{mishra2024finetuning} present significant advances through the use of fine-tuned models, such as Gemma-7B, to generate diagrams in Mermaid~\footnote{Mermaid --- \href{https://mermaid.js.org/}{https://mermaid.js.org/}}, their proposal is broad and generic, aimed at creating various types of diagrams. While it includes cases involving logical ER modeling, the study does not provide a dedicated analysis of this representation or investigate the conceptual interpretation challenges associated with transforming textual requirements into a logical schema. In contrast, the present work focuses exclusively on logical ER diagram generation, adopting a carefully developed approach for this purpose-from the creation of a specific dataset to the definition of structured evaluation criteria. Moreover, it proposes a flexible solution, not limited to a single trained model, allowing the use of different LLMs and prompt engineering strategies, thus enhancing its adaptability to different contexts and languages.

Finally, \citep{salem2024generating} presents a tool that combines LLMs with classic NLP techniques to extract information from textual requirements -- or even from images of ER diagrams -- and directly convert them into SQL commands. While the proposal is promising by focusing on automating the logical modeling stage, the work lacks clarity in fundamental methodological aspects. There are no precise descriptions of the internal workings of the tool, the criteria used for semantic extraction, or the specific role of LLMs in the process. This lack of information compromises reproducibility and hinders a critical evaluation of the approach's effectiveness. Furthermore, the tool does not include visual generation — an essential step for conceptual validation and effective communication between analysts and stakeholders. In this regard, the present work differentiates itself by offering a complete solution for conceptual database modeling, with a clear focus on the automatic and visual generation of logical relational diagrams from natural language, supported by a robust evaluation methodology, which includes model comparison and empirical validation with users.

\section{Methodology}
Three requirements documents were sequentially developed for the same factual context -- namely, the management of individuals accessing a hospital, within the context of human resources and security systems. The documents differ exclusively in the progressive increase in the complexity of the entity-relationship modeling, resulting from the cumulative addition of business constraints.

In order to simulate requisites provided by a non-expert client, the use of data modeling-specific jargon was deliberately avoided. This methodological choice underpins the empirical evaluation of the language model, based on the hypothesis that it is capable of acting as a specialist by identifying and structuring conceptual-to-logical models from semantically coherent specifications, even in the absence of explicit technical terminology.

The requirements documents were produced manually and subsequently reviewed with the assistance of a language model, used as a quality assurance tool to detect and minimize ambiguities and inconsistencies. Additionally, a reference logical database diagram corresponding to the most complex requisite document was created and used as a gold standard for comparison with the models generated by the evaluated system.

The experimental design comprises two independent factors. The first factor is the Requirements Scenario, represented by three requirement documents describing the same domain with progressively increasing complexity (Requirements Scenarios 1–3). The second factor is the Prompting Strategy, represented by three prompting configurations (One-shot, Chain-of-Thought, and Chain-of-Thought + Verifier). Each language model was evaluated under every combination of requirements scenario and prompting strategy.

\subsection{Main Processes}
\label{subsec:main-processes}

\begin{figure}
    \centering
    \includegraphics[width=0.7\columnwidth]{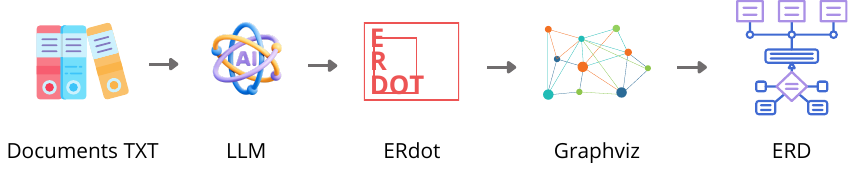}
    \caption{Flow diagram showing the main processes of the approach, from input to output, and their interactions.}
    \label{Fig1}
\end{figure}

The proposed workflow was implemented through a set of open-source artifacts designed to automate the transformation of textual requirements (.txt files) into logical relational diagrams. \textbf{Figure \ref{Fig1}} summarizes the overall process, which comprises three sequential stages:

\begin{itemize}
\item \textbf{Inference and Elicitation:} Requirements are processed by the selected LLM (Qwen-3 or GPT-5.1) through the Prompt Engineering Layer, which applies specific cognitive induction strategies (e.g., CoT, Verifier) to generate a conceptual model.
\item \textbf{Normalization:} The output is intercepted by a normalization script that performs cleaning, JSON parsing and validation, and basic structural checks on the expected fields and data types. This step ensures that the output has the required format for subsequent processing by ERDot.
\item \textbf{Automated Rendering:} The validated JSON is converted into a .dot file via ERDot~\footnote{ERDot --- \href{https://github.com/ehne/ERDot}{https://github.com/ehne/ERDot}} and subsequently rendered by Graphviz~\footnote{Graphviz --- \href{https://graphviz.org/}{https://graphviz.org/}} into PNG and SVG formats.
\end{itemize}

To ensure complete reproducibility, the source code, prompts, and experimental data are publicly available. The Normalization and Automated Rendering stages were implemented using the {visual-ER tool}~\footnote{visual-ER --- \href{https://github.com/EduardaFarias/visual-ER}{https://github.com/EduardaFarias/visual-ER}}. All artifacts generated throughout the experimental pipeline -- including the textual requirements, raw LLM outputs, normalized JSON representations, generated diagrams, the reference logical database diagram used as the gold standard, and the complete checklist-based analyses for every combination of requirement, prompting strategy, and language model -- are accessible in the \textit{llm-er-modeling} repository~\footnote{llm-er-modeling --- \href{https://github.com/ArthurHappx/llm-er-modeling}{https://github.com/ArthurHappx/llm-er-modeling}} repository.

\subsection{Evaluation with Different LLMs}

The experiments were conducted using two high-performance LLMs: Qwen-3 and GPT-5.1. The models were selected to represent two complementary categories of current LLMs: a high-performing open-weights model and a proprietary general-purpose model with strong reasoning capabilities. Importantly, the objective was not to select models specifically trained for database modeling, but to investigate whether state-of-the-art reasoning and technical capabilities transfer effectively to conceptual database modeling, which is a software engineering task that requires the interpretation of natural-language requirements and their transformation into a structured logical representation.

The selection was based on the following criteria:

\begin{itemize}
\item \textbf{Qwen-3}: represents the open-weights category and was selected for its strong performance in coding, mathematical reasoning, and instruction following. These capabilities are relevant to the task because database modeling requires multi-step reasoning over requirements, identification of entities and relationships, and adherence to a predefined structured output format.

\item \textbf{GPT-5.1}: represents a high-performing proprietary general-purpose model with advanced reasoning capabilities. Its inclusion provides a strong reference for evaluating whether a model with state-of-the-art general reasoning performance can reliably transform software requirements into consistent conceptual database structures.
\end{itemize}

This selection therefore allows the study to examine a relevant transfer question: whether strong general reasoning and software-engineering capabilities are sufficient to perform conceptual database modeling reliably. The comparison between an open-weights model and a high-performing proprietary model also provides complementary evidence regarding the extent to which the observed limitations are model-specific or reflect broader challenges in applying LLMs to this task.

\subsection{Experimental Framework for LLM Evaluation}

A reusable experimental framework was developed for the systematic execution of LLMs applied to logical representation. The infrastructure abstracts interaction with the models by encapsulating environment configuration, invocation, and response collection in a structured (JSON) format, ensuring experiment reproducibility and traceability.

\paragraph{Prompt Engineering Layer}
The Prompt Engineering layer constitutes the main source of experimental variation, enabling the isolation of the effects of different prompting strategies on the quality of logical schema. Three prompting strategies were evaluated, each representing a distinct experimental treatment:

\begin{itemize}
\item \textbf{One-shot (Baseline):} Direct instructions accompanied by a single input-output example, without an explicit intermediate reasoning or verification stage, serving as the baseline condition.
\item \textbf{Chain-of-Thought (CoT):} Prompts the model to follow an explicit five-step reasoning process before producing the logical ER model, covering key elements of logical ER modeling.
\item \textbf{CoT + Verifier:} Extends CoT with a three-stage pipeline: CoT generation, independent evaluation, and correction, using separate prompts for each stage.
\end{itemize}

Combined with the three Requirements Scenarios, these strategies result in nine experimental conditions for each evaluated language model.

\paragraph{Chain-of-Thought Flow}

In the CoT strategy, the model is prompted to produce, within a single response, an explicit five-step reasoning process before emitting the final JSON schema: (1) \textit{Entity Identification}, in which candidate entities are identified from the requirement text; (2) \textit{Attribute Determination}, in which attributes, primary keys, constraints, and data types are defined for each entity; (3) \textit{Relationship Identification}, in which associations between entities are identified and the possible need for intermediary tables is considered; (4) \textit{Cardinality and Foreign Key Determination}, in which cardinalities are determined using the ERDot notation ($1$: exactly one, $?$: zero or one, $*$: zero or more, and $+$: one or more), together with the corresponding foreign key placements; and (5) \textit{Schema Validation}, in which the model checks primary key uniqueness, foreign key validity, cardinality consistency, many-to-many handling, and the completeness of entities, attributes, and relationships before generating the final JSON schema.

\paragraph{CoT + Verifier Flow}

The CoT + Verifier strategy extends the previous flow with a separate validation pass performed through a dedicated verifier prompt applied to the same LLM~\footnote{The verifier was executed with a temperature of 0.2, while all other prompting strategies used a temperature of 0.0. A small non-zero temperature was used to introduce limited stochasticity in the verification pass when using the same LLM for generation and verification.}. The verifier receives both the original requirement and the schema generated by the CoT stage and critically inspects the schema against six predefined criteria: (1) primary key validation (exactly one per table), (2) foreign key validation (valid table and column references), (3) relationship cardinality validation, in which each relationship is examined to determine whether both cardinalities are logically consistent with the requirement and whether the foreign key is placed on the appropriate side of the relationship, (4) entity completeness (coverage of entities mentioned in the original requirement), (5) attribute completeness and data type appropriateness, and (6) correct handling of many-to-many relationships through intermediary tables. For each criterion, the verifier explicitly reports whether the check is satisfied and identifies the corresponding violations when it is not. The verifier produces a structured error report containing the identified issue, its impact, a suggested fix, and a confidence rating. The original schema, verification results, and requirement are then provided to the same LLM in a correction prompt, which generates a corrected schema using the five-step reasoning process. No additional verification cycle is performed after this correction pass.

\begin{table}[h]
\centering
\small
\caption{Summarized sample interaction for the CoT + Verifier strategy.}
\label{tab:sample-interaction}
\begin{tabular}{|p{1.6cm}|p{5.8cm}|}
\hline
\textbf{Stage} & \textbf{Content (condensed)} \\
\hline
Input & Requirement excerpt describing patients, appointments, and doctors. \\
\hline
CoT Step 1--4 & ENTITIES: [Patient, Doctor, Appointment]; ATTRIBUTES: Patient(*id, name, ...); RELATIONSHIPS: Patient--Appointment, Doctor--Appointment; CARDINALITY: Patient:id 1--* Appointment:patient\_id \\
\hline
CoT Step 5 + JSON & VALIDATION: all PKs present, FKs valid $\rightarrow$ JSON OUTPUT emitted \\
\hline
Verifier & PRIMARY KEY: VALID; FOREIGN KEY: Appointment.doctor\_id INVALID (references non-existent column) $\rightarrow$ ERROR \#1 reported with suggested fix \\
\hline
Correction & Corrected JSON regenerated with fixed foreign key reference, based on error report and original requirement \\
\hline
\end{tabular}
\end{table}

\subsection{ER Diagram Normalization and Generation Script}

Although the study is grounded on Entity–Relationship (ER) modeling principles, the diagrams generated in the experiments are not intended to represent pure conceptual ER models in the sense originally proposed by \citep{chen1976}. The rendering process relies on the ERDot tool, which produces logical relational diagrams enriched with implementation-oriented elements, such as primary and foreign keys, attribute data types, and relational constructs introduced for visualization purposes. Consequently, the generated artifacts should be interpreted as logical data models derived from conceptual modeling decisions, rather than strictly conceptual ER diagrams. This distinction is important when interpreting the evaluation criteria and the reported modeling limitations.

The operational flow is divided into three main stages:
\begin{itemize}
\item \textbf{Preprocessing and Cleaning:} The script extracts the JSON content from heterogeneous LLM outputs, handling formatting artifacts such as Markdown code blocks, escaped characters, and residual text surrounding the JSON object.

\item \textbf{JSON Validation and Structural Normalization:} The extracted content is parsed and validated as JSON. The script then checks whether the main elements required by ERDot are present in the expected form, including the collections representing entities and relationships and the diagram label. It also converts fields that may be returned in different representations into the format expected by ERDot. These checks concern only syntactic and structural compatibility and do not assess the semantic correctness of entities, attributes, relationships, or cardinalities.

\item \textbf{Automated Rendering:} The validated and normalized JSON representation is converted into the intermediate DOT format via ERDot and subsequently rendered into PNG and SVG files using Graphviz.
\end{itemize}

This automated component eliminates the need for manual intervention in diagram construction and provides a consistent visualization procedure across all experimental conditions, reducing potential variation arising from manual diagram rendering.

It is important to emphasize that the visualization serves only as preliminary support for model inspection, while the normalized JSON representation is consulted whenever necessary to verify the underlying model. Thus, potential issues arising from the graphical rendering are not attributed to the LLM-generated output. Consequently, the errors reported in the evaluation refer to the underlying normalized representation rather than to artifacts introduced during visualization.

\subsection{Qualitative Evaluation}
\label{subsec:qualitative-evaluation}

The diagrams generated were independently evaluated using a structured evaluation framework combining qualitative judgment with quantitative operationalization. The qualitative component was retained because the evaluation of conceptual-to-logical database models involves aspects that cannot be fully captured by direct structural comparison alone, particularly with respect to semantic adequacy, conceptual clarity, and the interpretation of business requirements.

The qualitative evaluation adopts a subjective classification procedure guided by explicit conceptual criteria, with the aim of characterizing different quality levels associated with the evolution of a logical relational diagram. The diagrams were organized into an ordinal ranking of four levels (L1--L4), representing a progression from minimal functional modeling to conceptually robust models prepared for scope evolution. These levels are described as follows.

Because the experimental pipeline produces logical relational diagrams through ERDot rather than pure conceptual ER diagrams, the evaluation combines classical conceptual modeling criteria (e.g., entity identification, relationship semantics, cardinalities, and completeness) with logical design aspects exposed by the generated artifacts (e.g., key definitions, mandatory attributes, and implementation-oriented constraints). Therefore, some evaluation items refer to logical schema quality rather than exclusively to conceptual ER modeling.

\paragraph{L1 -- Basic Scope Recognition:}
Defines the minimum level of modeling required for the schema to be conceptually functional. Essential domain entities are identified and represented. Minimum attributes necessary to characterize each entity are present. Direct and trivial relationships between entities are correctly established. A model that omits essential entities or relationships required for the basic operation of the described domain may fail to reach this level, even when some individual elements are correctly represented.

\paragraph{L2 -- Semantic Quality and Conceptual Clarity:} Corresponds to a modeling level capable of facilitating the understanding of the domain and the described requirements. Entity names are clear, descriptive, and unambiguous. Attribute names adequately reflect their meaning and content. Primary and foreign keys follow a consistent pattern. The model structure supports readability and conceptual maintainability.

\paragraph{L3 -- Structural Robustness and Data Integrity:} Encompasses models that demonstrate explicit concern for data integrity without relying on external mechanisms. Cardinalities and relationship types correctly represent business rules. Logical integrity constraints exposed by the generated schema are consistently represented (e.g., primary keys, foreign keys, mandatory attributes). The model exhibits greater independence from external validations or additional application logic.

\paragraph{L4 -- Extensibility and Model Evolution:} Represents conceptually mature models, prepared for natural expansion and evolution of the system scope. Anticipated domain changes can be incorporated with reduced structural impact. The model supports long-term maintenance and evolution. Structural organization contributes to query efficiency and higher-complexity requirements. \\

The evaluation was guided by a conceptual task list, used both to support the evaluators' qualitative judgment and to provide a basis for quantitative measurement (detailed in \textbf{Section~\ref{subsec:quantitative-operationalization}}). The same checklist therefore serves two complementary purposes: it preserves the qualitative interpretation of modeling quality while allowing the satisfaction of individual criteria to be systematically encoded and aggregated into quantitative scores.

\subsubsection{Task List}

The checklist comprises 26 items across six dimensions. The criteria used for each dimension are defined as follows.

\paragraph{Completeness}
(1) All domain entities have been identified. 
(2) All required relationships are present in the diagram. 
(3) There are sufficient attributes to characterize each entity. 
(4) Cardinalities correctly represent business rules. 
(5) The model covers all requirements identified in the scope.

\paragraph{Correctness}
(1) Attributes are associated with the correct entities. 
(2) Relationship types (1:1, 1:N, N:N) are correct. 
(3) There are no relationships not required by the domain (``spurious relationships''). 
(4) There are no duplicated or inconsistent attributes. 
(5) There are no unnecessary entities. 
(6) Minimum constraints (UNIQUE, NOT NULL) are correctly mapped.

\paragraph{Clarity}
(1) Entity names are clear, descriptive, and unambiguous. 
(2) Attribute names accurately reflect their content. 
(3) The layout supports readability (subareas separated by context). 
(4) Primary and foreign keys follow a consistent pattern. 
(5) There are no multiple different names for the same concept.

\paragraph{Simplicity}
(1) Absence of entities with an excessive number of attributes (e.g., > 12). 
(2) Absence of unnecessary complex relationships. 
(3) Decomposition of large logical ER diagrams into smaller subdiagrams. 
(4) Minimization of overly deep generalization/specialization hierarchies (e.g., > 3). 
(5) Absence of n-ary relationships where binary relationships would suffice.

\paragraph{Flexibility}
(1) Anticipated domain changes can be implemented without major impact. 
(2) Overly specific entities are avoided (preference for general concepts). 
(3) Subtypes are used only when necessary.

\paragraph{Implementability}
(1) There are no unnecessary complex triggers. 
(2) Attribute types are compatible with real data types.

\subsection{Quantitative Operationalization}
\label{subsec:quantitative-operationalization}

To complement the qualitative evaluation, the conceptual checklist was operationalized as a quantitative Checklist Quality Score (CQS). Each checklist item was encoded as a binary outcome. This operationalization allows the qualitative judgments applied during the evaluation to be aggregated into numerical scores while preserving the original conceptual dimensions of the assessment.

For a generated model $i$, the overall Checklist Quality Score is defined as

\begin{equation}
\mathrm{CQS}_{i}
=
\frac{\sum_{j=1}^{26} x_{ij}}{26}
\times 100,
\label{eq:cqs}
\end{equation}

\noindent where $x_{ij}=1$ when checklist item $j$ is satisfied by model $i$, and $x_{ij}=0$ otherwise. The resulting score ranges from 0 to 100, with higher values indicating a greater proportion of satisfied evaluation criteria.

Dimension-level scores were calculated independently for Completeness, Correctness, Clarity, Simplicity, Flexibility, and Implementability. For each dimension, the score represents the percentage of satisfied checklist items belonging to that dimension. These scores allow the analysis to distinguish between different aspects of model quality rather than relying exclusively on the aggregate CQS.

The CQS was calculated for every generated model and subsequently aggregated across the three requirement scenarios for each combination of LLM and prompting strategy. The mean CQS was used to characterize the overall performance of each experimental condition, while the sample standard deviation across the three requirement scenarios was used to characterize variation in performance across requirements of increasing complexity. The quantitative score is intended as a complementary measure rather than a replacement for the qualitative classification.

\begin{table*}[htbp]
    \centering
    \caption{CQS by quality dimension and overall ranking for each LLM--prompt combination.}
    \label{tab:cqs-ranking}
    \resizebox{\textwidth}{!}{%
        \begin{tabular}{llrrrrrrrrr}
            \toprule
            \textbf{Model} &
            \textbf{Prompt} &
            \textbf{Comp.} &
            \textbf{Corr.} &
            \textbf{Clar.} &
            \textbf{Simpl.} &
            \textbf{Flex.} &
            \textbf{Impl.} &
            \textbf{CQS Mean} &
            \textbf{SD} &
            \textbf{Rank} \\
            \midrule
            GPT-5.1 & CoT + Verifier & 33.33 & 50.00 & 93.33  & 86.67 & 77.78 & 50.00 & 65.38 & 11.54 & 1 \\
            GPT-5.1 & CoT            & 26.67 & 38.89 & 86.67  & 86.67 & 66.67 & 50.00 & 58.97 & 16.01 & 2 \\
            Qwen-3  & CoT + Verifier &  0.00 & 44.44 & 100.00 & 86.67 & 33.33 & 33.33 & 52.56 &  4.44 & 3 \\
            Qwen-3  & CoT            &  0.00 & 44.44 & 86.67  & 80.00 & 44.44 & 33.33 & 50.00 &  6.66 & 4 \\
            GPT-5.1 & One-shot       & 33.33 & 22.22 & 73.33  & 73.33 & 33.33 & 50.00 & 47.44 &  9.68 & 5 \\
            Qwen-3  & One-shot       &  6.67 & 33.33 & 60.00  & 86.67 & 55.56 & 50.00 & 47.44 &  5.88 & 5 \\
            \bottomrule
        \end{tabular}%
    }
    \begin{flushleft}
        \footnotesize
        \textit{Note.} Comp. = Completeness; Corr. = Correctness;
        Clar. = Clarity; Simpl. = Simplicity; Flex. = Flexibility;
        Impl. = Implementability. All scores are percentages.
    \end{flushleft}
\end{table*}

\section{Results}

The quantitative and qualitative evaluations provide complementary perspectives on the generated logical relational models. The quantitative analysis allows systematic comparison between models and prompting strategies, while the qualitative comments and the L1--L4 classification are used to assess whether the generated schemas reach meaningful levels of functional and structural adequacy.

Despite the relative differences observed among the experimental conditions, the overall results remain unsatisfactory for the modeling task investigated. No evaluated model--prompting configuration achieved complete coverage of the requirements: in particular, none of the generated models satisfied all Completeness criteria. This limitation is particularly important because Completeness concerns the identification of the domain entities, relationships, attributes, cardinalities, and requirements necessary to represent the specified system.

\textbf{Reading the Qualitative Results.} Due to space constraints, the complete textual requirements and raw LLM outputs cannot be reproduced in this paper. The examples presented in this section are therefore illustrative excerpts of the broader evaluation and should not be interpreted as an exhaustive comparison of all generated models. We strongly recommend that readers consult the complete requirement document alongside this section, since the qualitative observations are grounded in the specific constraints and concepts defined therein. For each requirement scenario, the \href{https://github.com/ArthurHappx/llm-er-modeling}{\textit{llm-er-modeling}} further provides the complete checklist-based evaluation for every model--prompting configuration, including evaluators' comments detailing the criteria used to identify and assess omissions, redundancies, incorrect relationships, cardinality errors, and other structural deficiencies. These artifacts allow the reader to examine the complete evidence underlying the qualitative observations

The quantitative results should therefore not be interpreted merely as evidence that one configuration is ``better'' than another. Rather, they reveal differences in the degree and nature of the deficiencies exhibited by the generated schemas. Some configurations obtained comparatively higher scores in dimensions such as Clarity and Simplicity while still failing to represent essential domain concepts. At the qualitative level, none of the analyzed results reached L3 (Structural Robustness and Data Integrity), and some generated models did not satisfy the minimum conditions required for L1 (Basic Scope Recognition), as they omitted essential entities or relationships required for a minimally functional representation of the described domain.

\subsection{Quantitative Checklist and Reference-Based Evaluation}
\label{sec:quantitative-evaluation}

\textbf{Table~\ref{tab:cqs-ranking}} presents the mean score for each checklist dimension and the overall CQS ranking. Scores are expressed as percentages, and the standard deviation of CQS is calculated across the three requirement scenarios. GPT-5.1 with CoT + Verifier obtained the highest mean CQS ($M=65.38$, $SD=11.54$), followed by GPT-5.1 with CoT ($M=58.97$, $SD=16.01$). Qwen-3 with CoT + Verifier ranked third ($M=52.56$, $SD=4.44$), while Qwen-3 with CoT obtained a mean CQS of 50.00 ($SD=6.66$). Both one-shot conditions obtained the same overall mean CQS of 47.44, although Qwen-3 showed lower variation across the three scenarios.

The most critical result concerns Completeness. GPT-5.1 obtained its highest Completeness score with CoT + Verifier, but the mean score remained only 33.33. Qwen-3 obtained 0.00 in Completeness under both CoT configurations. Thus, even the comparatively stronger configurations failed to provide complete coverage of the requirements.

The contrast between Completeness and Clarity is particularly revealing. Qwen-3 with CoT + Verifier achieved 100.00 in Clarity and 86.67 in Simplicity while obtaining 0.00 in Completeness. This result demonstrates that a generated schema may be clear and structurally simple while still failing to represent essential elements of the specified domain. Consequently, organizational quality cannot be treated as evidence of adequate requirement coverage.

GPT-5.1 showed a more favorable profile in Completeness and Flexibility than Qwen-3, particularly when the verifier was introduced. Nevertheless, its highest Completeness score remained only 33.33. The quantitative evidence therefore supports the conclusion that the principal limitation of the evaluated systems is not merely presentation quality, but the ability to systematically identify and represent the concepts and constraints required by the specification.

Descriptively, GPT-5.1 obtained a 7.26-point higher mean CQS than Qwen-3, while CoT improved the mean CQS by 7.05 points over One-shot. Adding the Verifier produced a further 4.49-point increase over CoT. None of the prompting configurations achieved complete requirement coverage. These results suggest that additional reasoning and verification can improve some evaluated quality dimensions, but do not consistently resolve the fundamental modeling deficiencies identified in the generated schemas. However, none of these differences was statistically significant. It is important to reiterate that the primary objective of this study is not to exhaustively define or compare prompting techniques, but to investigate the general capability of LLMs to perform logical ER modeling. Given the exploratory nature of the study, these results should be interpreted as descriptive evidence rather than evidence of definitive superiority between models or prompting strategies. 

\subsubsection{Analysis of GPT-5.1}

GPT-5.1 achieved the highest quantitative performance across the evaluated configurations. Its highest mean CQS was obtained with CoT + Verifier (65.38). Nevertheless, as mentioned, the Completeness score remained low in both configurations where GPT-5.1 achieved its highest Completeness, indicating that substantial portions of the requirements were still not represented.

The qualitative analysis, in addition, reveals a systematic tendency to include entities, attributes, and relationships that were not requested in the original requirements. Although some additions are contextually plausible, they lack a reasonable criterion of necessity, resulting in redundancies and risks to data integrity.

\paragraph{Structural Redundancy and Integrity Risks}

The model frequently introduces relationships that could be derived through inference, compromising the simplicity of the schema. A clear example occurs in the hospital access situation:

Example of modeling via CoT (Chain of Thought):
\begin{quote}
\texttt{``Hospital:hospital\_id 1--* HospitalDepartment:hospital\_id'',}\\ \\
\texttt{``Hospital:hospital\_id 1--* VisitorAccess'',}\\ \\
\texttt{``HospitalDepartment:hospital\_department\_id 1--* VisitorAccess''}
\end{quote}

In this structure, the direct relationship between Hospital and VisitorAccess is redundant, since each access record is already linked to a Department, which in turn belongs to the Hospital. This configuration ignores transitive dependency and introduces a risk of inconsistency: a database derived from this model would allow recording an access for hospital A while referencing a department belonging to hospital B. Mitigating this issue would require additional, unnecessary constraints, thereby increasing implementation complexity.

The model also exhibits errors in relationship cardinalities. For example, it generated the relationship \texttt{Employee 1--* Physician}, although the requirements do not indicate that an employee may be associated with multiple physician records. Instead, the specification states that ``A single person may assume more than one role (for example: a physician may be admitted as a patient outside their working hours)'' and ``A nurse cannot be recorded as a physician (roles are distinct)'' while also explicitly establishing that roles such as nurse and physician are distinct. This describes the possibility of a person assuming different roles, rather than multiple instances of the same role being associated with one employee. The same type of cardinality error was observed in other relationships, including \texttt{Employee 1--* Nurse}, \texttt{Person 1--* Patient}, \texttt{Person 1--* Employee}, and \texttt{Patient 1--* EmployeeDependent}. Such incorrect cardinalities can directly affect the integrity of a logical schema by allowing multiple role or dependency associations where the requirements specify a one-to-one correspondence.

\paragraph{Over-specification and Undue ``Creativity''}

Furthermore, the formulation of the PersonRole entity for Requirement Scenario 1 is not requested in the document, since the roles performed by each individual within the scope are clearly delimited:
\begin{quote}
``The hospital receives individuals from various categories, some are employees, others are not. These include visitor, patient, physician, nurse, resident, researcher, and administrative staff''
\end{quote}
and the specification defines attributes for each of these roles, supporting their treatment as distinct entities within the scope of the requirements. While this addition may be useful depending on what is envisioned as a natural evolution of the defined system, the model fails to guarantee the integrity required within the delimited scope itself and therefore should not rely on creativity to introduce additional elements when the specified requirements have not yet been properly satisfied. \\

From the perspective of classical database theory, this behavior of GPT-5.1 undermines redundancy elimination and the correct representation of functional dependencies. It demonstrates difficulty in distinguishing fundamental attributes (essential to the entity) from derived attributes (which can be obtained through queries), resulting in structurally unstable schemas that fail to provide a solid foundation for data persistence.

\begin{table*}[ht]
\caption{Mean token usage and estimated inference cost for each model and prompting technique.}
\centering
\begin{tabular*}{\textwidth}{@{\extracolsep{\fill}}llcccc@{}}
\hline \hline
\textbf{Model} &
\textbf{Prompting Technique} &
\textbf{Mean Input Tokens} &
\textbf{Mean Output Tokens} &
\textbf{Mean Cost} &
\textbf{SD Cost} \\
\hline
GPT-5.1 & One-shot       & 2,854.000  & 2,860.667   & \$0.032 & \$0.004 \\
Qwen-3  & One-shot       & 2,892.000  & 3,621.000   & --      & --      \\
GPT-5.1 & CoT            & 2,608.000  & 8,560.667   & \$0.089 & \$0.017 \\
Qwen-3  & CoT            & 2,612.000  & 4,000.667   & --      & --      \\
GPT-5.1 & CoT + Verifier & 13,597.000 & 21,793.000  & \$0.235 & \$0.037 \\
Qwen-3  & CoT + Verifier & 9,579.667  & 10,764.667  & --      & --      \\
\hline \hline
\end{tabular*}
\label{tab:token_cost}
\end{table*}

\subsubsection{Analysis of Qwen-3}

Qwen-3 exhibited substantially lower Completeness than GPT-5.1. Its Completeness score was 0.00 under both CoT configurations and only 6.67 with One-shot. The qualitative evaluation indicates that these scores result primarily from the omission or inadequate representation of essential domain concepts.

An example of these failures appears in the modeling response for Requirement Scenario 1, submitted under the CoT with Verifier technique.

Regarding entity identification, the model represented all described roles as a single \textit{Employee} entity, omitting the \textit{Physician} and \textit{Nurse}, resulting in the loss of the conceptual separation required for role-specific attributes such as \textit{crm} and \textit{coren}, which apply respectively to physicians and nurses.

\begin{quote}
   ``Healthcare employees must have their professional credentials recorded (e.g., CRM for physicians and COREN for nurses).''
\end{quote}

In several cases, the model demonstrated difficulties in mapping relationships, particularly in situations that could be directly resolved through the explicit definition of foreign keys.

A lack of consistency was also identified in the strategy adopted to represent relationships: at times the relationship was explicitly modeled as such, while in other instances it was replaced merely by attributes suggestive of foreign keys, without conceptual uniformity. Additionally, the model repeatedly showed confusion in distinguishing between binary and n-ary relationships, directly affecting the structural and semantic clarity of the generated diagrams.

This issue can be observed, when addressing the requirement:

\begin{quote}
   ``Because patients and visitors do not have personalized cards, a card’s identifier number may be temporarily linked to the person’s record while in use.''
\end{quote}

The produced representation -- 
\begin{quote}
\texttt{IdentificationCard:card\_id ?--1 Visitor:card\_id}
\end{quote}
-- reveals an inadequate modeling of the temporary linkage described in the specification. In other words, the correct interpretation would be that a single card may be associated with more than one visitor, although not simultaneously. This flaw undermines the traceability intended by the fictional hospital network’s security team, which should be able to 
\begin{quote}
``Query access history by person, date, department and type (employee, visitor, patient).''
\end{quote}

\subsubsection{Direct Comparison Between the Models}

The two LLMs exhibited complementary failure patterns. Thus, the main distinction between the models is not the absence of modeling errors, but their nature: GPT-5.1 tends toward over-specification, whereas Qwen-3 tends toward under-specification. Neither behavior resulted in models that satisfied the requirements completely or reached the L3 level of structural robustness and data integrity.

\subsection{Cost and Efficiency Analysis}

\textbf{Table~\ref{tab:token_cost}} reports the mean input and output tokens and the estimated inference cost~\footnote{For GPT-5.1, the API cost was estimated using \$1.25 per million input tokens and \$10.00 per million output tokens. Energy and water consumption were also not measured in this study. Readers interested in estimating these costs and environmental impacts may consult tools such as the AI Energy Calculator provided by 1ClickImpact (\url{https://1clickimpact.com/ai-energy-calculator}). A detailed analysis of these aspects was not included due to the page limit.} for each model and prompting configuration. For CoT + Verifier, token usage includes all stages of the procedure: the initial CoT generation, verifier analysis, and post-verifier generation of the final response.

For GPT-5.1, the estimated cost increased from \$0.032 with One-shot to \$0.089 with CoT and \$0.235 with CoT + Verifier. Thus, CoT required approximately 2.8 times the cost of One-shot, while CoT + Verifier required approximately 7.3 times more. Although CoT + Verifier achieved the highest mean CQS (65.38), its Completeness remained limited to 33.33, and not reached L3. The remaining omissions and structural inconsistencies therefore prevent the generated models from being considered suitable for production use. Consequently, the substantial additional cost of the verification pipeline did not result in a big improvement in structural agreement with the reference.

For Qwen-3, monetary inference costs were not included in the analysis. As an open-weight model, Qwen-3 can be deployed on local infrastructure, in which case the associated costs depend on the computational resources and deployment configuration. Nevertheless, the token counts show the same general pattern: CoT + Verifier required substantially more tokens than either One-shot or CoT. Overall, the results indicate that increasing reasoning and verification substantially increases computational requirements, while the resulting quality improvements remain limited and do not guarantee complete requirement coverage or structural robustness.

\section{Conclusion and Future Work}

This study investigated the ability of large language models to transform textual requirements into logical relational database models under different prompting strategies. The results indicate that, although the evaluated models and prompting configurations exhibit different strengths and failure patterns, they do not yet provide sufficient reliability for autonomous end-to-end database modeling. In particular, the persistent difficulties in identifying essential entities and relationships, representing constraints consistently, and achieving complete requirement coverage limit the practical adequacy of the generated models.

The qualitative and quantitative evaluations also show that improvements in selected quality dimensions do not necessarily translate into functionally adequate models. Additional reasoning and verification can improve some aspects of the generated representations, but they do not consistently resolve the fundamental modeling deficiencies observed in the experiments. Consequently, the results support the use of LLMs as assistants to database modeling rather than as substitutes for the analysis and validation performed by domain and database specialists.

Among the contributions of this work are the construction of a dataset containing textual requirements and corresponding reference logical models, the implementation of a reproducible generation and evaluation pipeline, and the comparison of two LLMs under three prompting strategies. The combination of quantitative checklist-based evaluation and qualitative analysis provides a broader characterization of the capabilities and limitations of LLMs in conceptual-to-logical database modeling.

As future work, we plan to investigate LLMs as support tools for specific stages of the modeling process, such as requirement ambiguity detection, entity and relationship identification, constraint validation, and post-generation schema review. Rather than asking an LLM to perform the entire logical ER modeling task, future studies will investigate whether decomposing the process into smaller, specialized tasks, handled by agents dedicated to specific modeling stages leads to more complete and structurally adequate schemas. Future experiments should also consider larger and more diverse requirement benchmarks, covering different application domains and levels of complexity, to assess the generality of the observed findings and examine whether targeted verification mechanisms can improve requirement coverage and structural adequacy.

\bibliographystyle{apalike}
\bibliography{references}

\end{document}